# SU-ESRGAN: Semantic and Uncertainty-Aware ESRGAN for Super-Resolution of Satellite and Drone Imagery with Fine-Tuning for Cross Domain Evaluation

[1]Prerana Ramkumar
College of Engineering,
American University of Sharjah,
Sharjah, United Arab Emirates
g00100339@aus.edu

*Abstract*—Generative Adversarial Networks (GANs) have achieved realistic super-resolution (SR) of images however, they lack semantic consistency and per-pixel confidence, limiting their credibility in critical remote sensing applications such as disaster response, urban planning and agriculture. This paper introduces *Semantic and Uncertainty-Aware ESRGAN (SU-ESRGAN)*, the first SR framework designed for satellite imagery to integrate the ESRGAN, segmentation loss via DeepLabv3 for class detail preservation and Monte Carlo dropout to produce pixel-wise uncertainty maps. The SU-ESRGAN produces results (PSNR, SSIM, LPIPS) comparable to the Baseline ESRGAN on aerial imagery. This novel model is valuable in satellite systems or UAVs that use wide field-of-view (FoV) cameras, trading off spatial resolution for coverage. The modular design allows integration in UAV data pipelines for on-board or post-processing SR to enhance imagery resulting due to motion blur, compression and sensor limitations. Further, the model is fine-tuned to evaluate its performance on cross domain applications. The tests are conducted on two drone based datasets which differ in altitude and imaging perspective. Performance evaluation of the fine-tuned models show a stronger adaptation to the Aerial Maritime Drone Dataset, whose imaging characteristics align with the training data, highlighting the importance of domain-aware training in SR-applications.

## I. Introduction

High resolution satellite and UAV imagery is important for urban planning, precision agriculture, defense, intelligence and climate change research. Many factors such as high altitude and field of view, sensor limitations, atmospheric conditions, motion blur and compression produce low resolution imagery. Traditional CNN-based SR models optimized for pixel fidelity (SRCNN [1], EDSR [2], RCAN [3]) achieve high PSNR but smooth outputs that lack finer details. GANs such as SRGAN [4] and ESRGAN [5] improve quality by adding adversarial and perceptual loses, however, these lack semantic consistency and uncertainty estimates crucial for geospatial analysis. Image SR can be used for unauthorized surveillance, misidentification, or the creation of deceptive content. Therefore, semantic consistency which ensures that meaningful features are preserved and uncertainty mapping, that provides pixel wise confidence are vital for reliable analysis and decision making. The contributions made by this paper are formalized below.

This paper proposes **Semantic and Uncertainty-Aware ESRGAN (SU-ESRGAN),** a novel SR framework that combines semantic segmentation and Bayesian uncertainty estimation into the ESRGAN. SU-ESRGAN augments the existing adversarial-perceptual loses with segmentation loss (using DeepLabv3 [6]) for class and object preservation, and utilizes Monte Carlo Dropout [7] at test time for the production of pixel-wise uncertainty estimation. Trained on UCMerced Land Use [8] and AID [9] datasets, the SU-ESRGAN achieves performance comparable to the baseline ESRGAN model with indifferent training time consumption. The SU-ESRGAN is further fine-tuned for two drone based datasets: UAVid [10] and Aerial Maritime Drone Dataset [11] to evaluate cross-domain performance and the potential use of the model in drone based applications. The cross-domain fine-tuning demonstrates that domain-aware training impacts adaptation performance. This establishes benchmarks for satellite-to-drone domain transfer in super-resolution. The paper introduces a modular design that enables deployment of uncertainty-aware super-resolution in operational satellite and UAV systems. This framework makes reliable SR accessible for geospatial analysis with seamless integration into existing remote sensing pipelines.

## II. Related Work

### A. Single-Image Super-Resolution and GANs

Deep SR models have evolved from early models from SRCNN to advanced architectures such as EDSR and RCAN, which are optimized for high PSNR and produce excessively smoothed outputs. SRGAN and ESRGAN improve SR realism using adversarial loss and perceptual loss with ESRGAN introducing Residual in Residual Dense Blocks and a relativistic discriminator. These approaches perform image enhancement but hallucinate features due to the lack of semantic based recognition of classes in imagery.

### B. Semantic Guidance In Super-Resolution

Recent work such as Semantic Segmentation Guided (SSG-RWSR [12]) optimizes SR with a segmentation loss, using a segmentation network to steer reconstruction toward semantic fidelity. DeepLabv3 [6], a state of the art segmentation model, uses atrous convolution and Atrous Spatial Pyramid Pooling (ASPP) that is effective for extracting multi-scale semantic features. Leveraging DeepLabv3 in SR encourages outputs to remain consistent with scene classes, supporting semantic consistency which is an important property for human perception in SR.

### C. Uncertainty in Deep Learning for SR

Test-time dropout (Monte Carlo dropout) provides a practical means to estimate predictive uncertainty in deep networks. Recent work applies MC dropout to generate uncertainty maps in SR, highlighting unreliable regions and supporting confidence assessment in SR outputs [13]. Our work adopts MC-dropout in the SR generator to yield pixel-wise variances alongside the SR image, enabling risk-aware interpretation.

### D. Remote Sensing Context and Domain Adaptation

High-resolution remote sensing images are vital for applications such as land-use mapping and disaster management. However, sensors often trade spatial resolution for coverage, and domain gaps between satellite and drone imagery complicate SR tasks. Tang *et al* recently combined SR with domain adaptation for segmentation [14]. The paper noted that sensors produce large resolution disparities (a car at 4m vs 1m resolution appears very different). This motivates domain-aware evaluation: this paper presents a fine-tuned SU-ESRGAN on drone imagery with different altitude and view to assess adaptation.

## III. METHODOLOGY

The SU-ESRGAN extends ESRGAN by integrating (1) a Bayesian uncertainty via Monte Carlo dropout and (2) semantic guidance via DeepLabv3 segmentation loss. The architecture comprises of the following:

### A. Generator Architecture

The SU-ESRGAN's generator integrates Monte Carlo Dropout for uncertainty estimation on the RRDB backbone. First, a convolutional layer maps the 3 channel low-resolution input into 64 feature maps. Next, a dropout layer (p=0.2) is used to inject stochasticity for Bayesian inference. This output is passed through 5 RRDBs to enhance details and aggregated to 64 channels through a 3x3 convolution. These aggregated features are added to the early dropped features to preserve fine details. Upsampling is performed in two states, where each consists of a 3x3 convolution to 256 channels, PixelShuffle x 2 and LeakyReLU(0.2). The upsampling achieves 4x spatial enlargement. Finally, a convolutional layer reduces channels from 64 to three, producing the SR RGB output.

The discriminator remains unchanged from the ESRGAN's discriminator design.

### B. Semantic Guidance

Traditional SR methods primarily rely on Mean Squared Error (MSE) and Peak-Signal-To-Noise Ratio (PSNR) which leads to overly smooth or blurry outputs. Semantic consistency loss is calculated to ensure that the super-resolved outputs preserve the same semantic layout as the ground-truth image. This loss is calculated as the average L1 distance between per-pixel class assignments produced by DeepLabv3. Specifically, if $S_{SR}^{(i)}$ and $S_{HR}^{(i)}$ are the arg-max class indices at pixel *i* for the SR and HR images, respectively, then:

$$L_{sem} = -\frac{1}{N}\sum_{i=1}^{N} |S_{SR}^{(i)} - S_{HR}^{(i)}| \qquad (1)$$

Minimizing this loss allows the SR network to produce outputs with semantic segmentation maps similar to the HR reference, preserving class-level consistency.

### C. Inference and Uncertainty Estimation

The Monte Carlo dropout mechanism measures uncertainty by keeping dropout layers active throughout inference. The dropout layers are kept in training model during test, allowing for stochastic behavior over several forward passes. T = 20 stochastic forward passes over the network mare made via the uncertainty estimation procedure, producing T distinct super-resolved predictions for every input. The final outputs include the per-pixel standard deviation σ, which is the square root of the variation across forecasts, and the per-pixel mean μ, which is the average of all T predictions. Higher values on the standard deviation map indicate areas where the model has less confidence in its predictions.

$$\mu = \frac{1}{20}\sum_{t=1}^{20} SR_t \, , \, \sigma = \sqrt{\frac{1}{T}\sum_{t=1}^{20}(SR_t - \mu)} \qquad (2)$$

### D. Domain Adaptation Strategy

To evaluate cross-domain generalization capabilities, the trained SU-ESRGAN model is fine-tuned on drone datasets including UAVid and Aerial Maritime Drone Dataset. To prevent overfitting, the fine tuning process uses a slower learning rate of $1 \times 10^{-5}$ and short training schedules. The domain adaptability is assessed by comparing PSNR, LPIPS and SSIM metrics against the base model performance. This evaluation strategy shows the model's ability to transfer knowledge across different imaging platforms and conditions.

## IV. EXPERIMENT

### A. Datasets and Data Preprocessing

The datasets used to train the baseline ESRGAN and SU-ESRGAN are UC Merced LandUse (2100 images) and AID dataset (10,000 images) obtained from USGS National Map Urban Area Imagery collection and Google Earth respectively. The combined collection was divided into 7744 training (64%), 1936 validation (16%) and 2420 test (20%) images. The split was done in a way to preserve the original class distribution in the dataset. The split was followed by a round of bicubic interpolation to obtain corresponding low resolution (64x64) for the high resolution imagery (256x256) in each of the train, test and validation directories. The UAVid and Aerial Maritime Drone Datasets were split similarly, 60/20/20 for train, test and validation.

### B. Training

All training was performed on NVIDIA Tesla P100 GPU accessed via the machine learning platform, Kaggle [15]. Training loops were initially set up for 30 epochs with early stopping (patience=10) to prevent overfitting. The model from the best epoch was saved and loaded for evaluation. More details about the training are present in table 1.

### C. Evaluation Metrics

Each model was evaluated against a set of metrics using test images from its respective training dataset.

*a) PSNR (Peak Signal-To-Noise Ratio):* A pixel wise metric calculating image reconstruction quality based on the ratio of maximum signal power to noise power.

*b) SSIM (Structual Similarity Index Measure):* A perception based metric that assesses image similarity by considering luminance, contrast and structural information, providing a correlation with human visual perception.

*c) LPIPS (Learned Perceptual Image Patch Similarity):* Measures perceptual similarity by comparing high level-feature representations extracted from images using a pre-trained CNN.

*d) FID (Fréchet Inception Distance):* Quantifies the statistical similarity between the feature distributions of generated and real image sets.

The values obtained are formalized in table 2.

## V. Results and Analysis

The evaluation compared the Baseline ESRGAN, SU-ESRGAN and the fine-tuned versions. The output imagery obtained from the Baseline and SU-ESRGAN models are present in fig. 1 and fig. 2 respectively. The results achieved by the Baseline ESRGAN (PSNR: 25.99 dB, SSIM: 0.696, LPIPS: 0.2672, FID: 68.617) shall be considered as the benchmark for the rest of the analysis. The SU-ESRGAN variant, exhibited lower pixel-wise (PSNR: 25.01) and perceptual similarity (LPIPS: 0.3172) compared to the baseline, with a slightly higher FID. A notable observation was that SU-ESRGAN consistently produced visually blurrier results compared to the Baseline ESRGAN as shown in fig. 2. This suggests that the integration of segmentation loss has shifted the model's outputs towards prioritizing fidelity over the sharp, hallucinated details of the benchmark model.

Fine-tuning on the domain mismatched, UAVid dataset results in lower PSNR/SSIM and an unusually low FID of 3.750 suggesting mode collapse in generated images. Conversely, fine-tuning on the Aerial Maritime Drone Dataset with a better domain match yielded improved perceptual quality (SSIM: 0.742, LPIPS:0.1769) as seen in fig 3. This configuration resulted in a higher FID, indicating that while the model successfully generated perceptually superior and more diverse images, its overall distribution of generated images still deviates from the real maritime drone data. The uncertainty maps use color to visualize the model's confidence in the generated images. Bright colors such as white or yellow show high ambiguity. This especially occurs in areas of fine details. Darker colors such as black show low uncertainty, signifying consistent predictions in less ambiguous regions.

## VI. Conclusion

This paper designed and evaluated SU-ESRGAN, a novel SR model alongside the Baseline ESRGAN and fine-tuned configurations. Our design produced uncertainty mappings allowing for responsible analysis of SR images. Our findings indicated that SU-ESRGAN produced blurrier outputs compared to the baseline due to perceptual-distortion trade-off. The fine-tuning results emphasize that low FID, should be carefully interpreted as a potential indicator of model collapse rather than superior generation. Future work shall focus on improving SU-ESRGAN's design to mitigate blurriness in generated imagery along with emphasis on domain aware adaptation strategies for drone applications.

## Acknowledgment

P. Ramkumar thanks her parents, Smitha and Ramkumar, for their unwavering support and encouragement. She also acknowledges S. Bharadwaj for his valuable assistance and thoughtful discussions during the preparation of this work.

## References

[1] C. Dong, C. C. Loy, K. He, and X. Tang, "Image super-resolution using deep convolutional networks," *IEEE Trans. Pattern Anal. Mach. Intell.*, vol. 38, no. 2, pp. 295–307, Feb. 2016.

[2] B. Lim, S. Son, H. Kim, S. Nah, and K. M. Lee, "Enhanced deep residual networks for single image super-resolution," in *Proc. IEEE Conf. Comput. Vis. Pattern Recognit. Workshops*, 2017, pp. 1132–1140.

[3] Y. Zhang, K. Li, K. Li, L. Wang, B. Zhong, and Y. Fu, "Image super-resolution using very deep residual channel attention networks," in *Proc. Eur. Conf. Comput. Vis.*, 2018, pp. 286–301.

[4] C. Ledig *et al.*, "Photo-realistic single image super-resolution using a generative adversarial network," in *Proc. IEEE Conf. Comput. Vis. Pattern Recognit.*, 2017, pp. 4681–4690.

[5] X. Wang *et al.*, "ESRGAN: Enhanced super-resolution generative adversarial networks," in *Proc. Eur. Conf. Comput. Vis. Workshops*, 2018, pp. 63–79.

[6] L. Chen, G. Papandreou, I. Kokkinos, K. Murphy, and A. L. Yuille, "DeepLab: Semantic image segmentation with deep convolutional nets, atrous convolution, and fully connected CRFs," *IEEE Transactions on Pattern Analysis and Machine Intelligence*, vol. 40, no. 4, pp. 834-848, Apr. 2018, doi: 10.1109/TPAMI.2017.2699184.

[7] Y. Gal and Z. Ghahramani, "Dropout as a bayesian approximation: Representing model uncertainty in deep learning," in *Proc. Int. Conf. Mach. Learn.*, 2016, pp. 1050–1059.

[8] Y. Yang and S. Newsam, "Bag-of-visual-words and spatial extensions for land-use classification," in *Proc. ACM SIGSPATIAL Int. Conf. Adv. Geogr. Inf. Syst.*, 2010, pp. 270–279.

[9] G.-S. Xia *et al.*, "AID: A benchmark data set for performance evaluation of aerial scene classification," *IEEE Trans. Geosci. Remote Sens.*, vol. 55, no. 7, pp. 3965–3981, Jul. 2017.

[10] Y. Lyu, G. Vosselman, G.-S. Xia, A. Yilmaz, and M. Y. Yang, "UAVid: A semantic segmentation dataset for UAV imagery," *ISPRS J. Photogramm. Remote Sens.*, vol. 165, pp. 108–119, Jul. 2020.

[11] Roboflow, "Aerial maritime drone dataset," Roboflow Universe, 2020. [Online]. Available: https://universe.roboflow.com/roboflow-jvuqo/aerial-maritime. [Accessed: Dec. 15, 2024].

[12] A. Åkerberg, A. S. Johansen, K. Nasrollahi, and T. B. Moeslund, "Semantic segmentation guided real-world super-resolution," in *Proc. IEEE/CVF Winter Conf. Appl. Comput. Vis. Workshops*, 2022, pp. 449–458.

[13] M. S. Adapa, M. Zullich, and M. Valdenegro-Toro, "Uncertainty estimation for super-resolution using ESRGAN," *arXiv preprint arXiv:2412.15439*, 2024.

[14] Z. Tang, Y. Gao, Y. Xie, J. Zhu, Z. Wang, and J. Li, "Super-resolution domain adaptation networks for semantic segmentation via pixel and output-level aligning," *Front. Earth Sci.*, vol. 10, p. 974325, Aug. 2022.

[15] Kaggle Inc., *Kaggle: Your Machine Learning and Data Science Community*. [Online]. Available: https://www.kaggle.com/

TABLE I. Training Details For All Models

| Model | Training Images | Early Stopping (Patience = 10) | Time Consumption |
|---|---|---|---|
| Baseline ESRGAN | 7744 | 26 | ~3 hour |
| SU-ESRGAN | 7744 | 15 | ~3.5 hour |
| Finetuning – UAVid Dataset | 600 | 13 | ~0.35 hour |
| Finetuning – Aerial Maritime Drone | 372 | 15 | ~0.19 hour |

TABLE II. Evaluation Of Models

| Model | PSNR (Decibels) | SSIM | LPIPS | FID |
|---|---|---|---|---|
| Baseline ESRGAN | 25.99 | 0.696 | 0.2672 | 68.617 |
| SU-ESRGAN | 25.01 | 0.696 | 0.3172 | 75.230 |
| Finetuning – UAVid Dataset | 23.78 | 0.637 | 0.2448 | 3.750 |
| Finetuning – Aerial Maritime Drone | 24.77 | 0.742 | 0.1769 | 154.153 |

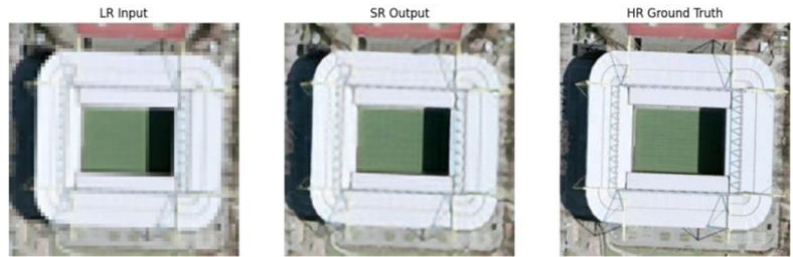

Fig. 1. LR input, HR ground truth and SR output produced by the Baseline ESRGAN

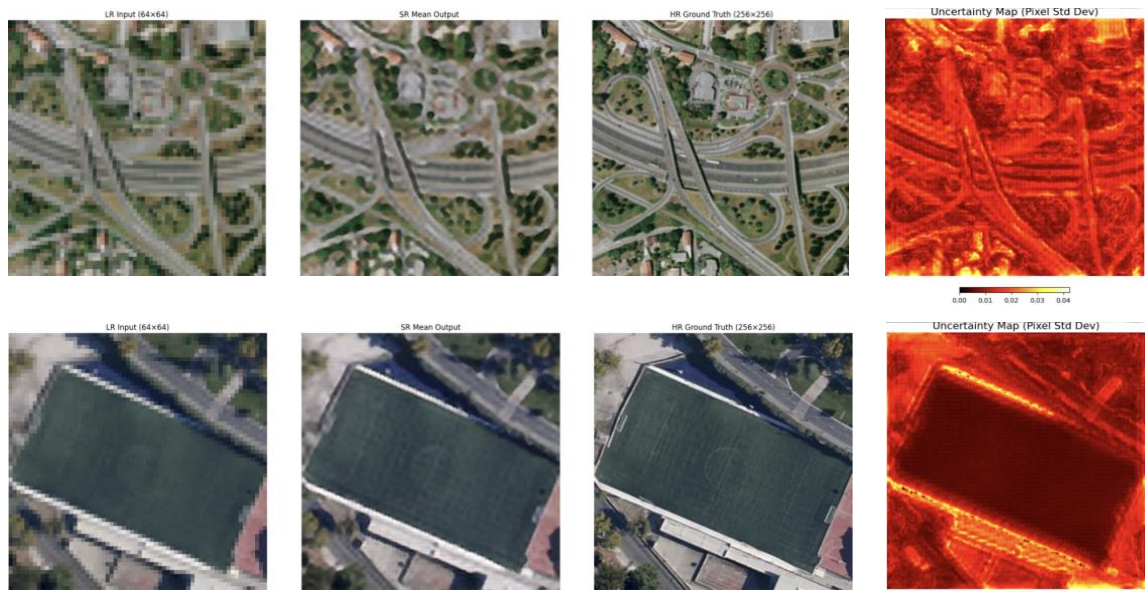

Fig. 2. LR input, SR output, HR ground truth and uncertainty mapping produced by the SU-ESRGAN on two different images

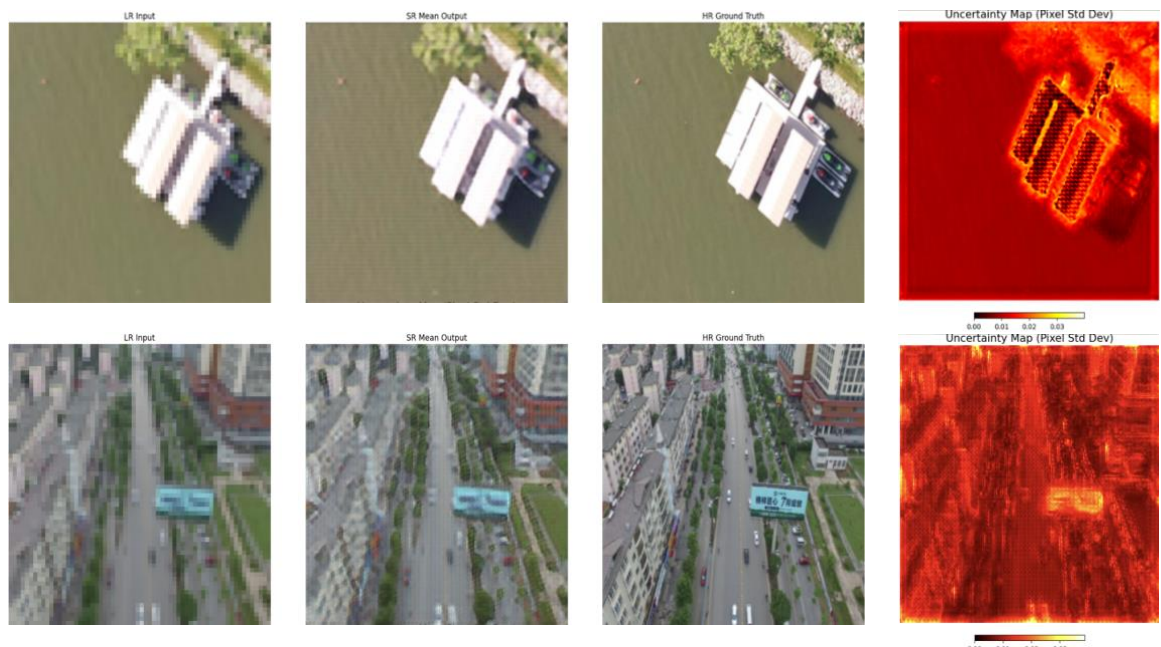

Fig. 3. Top – Fine-tuning on Aerial Maritime Dataset, Bottom – Fine-tuning on UAVid Dataset